# Dimensionality in Satisfaction Ratings

*Validating LLM annotation as a method to recover decomposed satisfaction across five axes of customer experience.*

Andrew Hong Jason Potteiger

Dimension Labs

## Abstract

We used a large language model (GPT-4.1) to annotate the text of ~9,000 support conversations at a global consumer-goods firm, decomposing customer-care satisfaction into component axes (overall, agent, outcome, product, and customer effort), and validated the LLM annotations against the satisfaction ratings customers gave themselves. Four of five axes track self-reported satisfaction closely (overall, agent, and outcome near an unadjusted 0.65; effort −0.54), while product satisfaction is weak against the available proxy. The unadjusted correlation also understates the alignment: the disagreements concentrate in a small, readable tail of divergent sessions rather than in general drift, and the overall correlation rises to 0.811 when only the severe divergences are excluded and to 0.914 when the full divergent tail is excluded. The axes are also highly collinear, and adding them to the overall score does not improve prediction of the customer's rating: the decomposition's value is not incremental prediction but attribution and coverage. And, with greater coverage the picture of the data changes. Read on every contact rather than the few that return a survey, satisfaction is markedly lower than the survey reports (a full-census 2.91 against the surveyed 3.62 on a five-point scale). The promise of decomposed satisfaction as a methodology is the ability to identify more nuanced drivers of customer experience in conversational data.

**Findings at a glance.** Each row names the cohort behind the number; sections give the full evidence.

| Finding | Headline number | Cohort | Section |
|---|---|---|---|
| The annotation tracks the customer's own rating | ρ = 0.65 unadjusted; 0.811 / 0.914 with severe / all divergences excluded | In-chat base (n = 385) | §§4, 7 |
| The component axes add no predictive power | Inter-axis ρ of 0.84–0.94; composites land within 0.02 of the overall axis alone | Collinearity sample (3,310) | §5 |
| The survey overstates satisfaction | Surveyed sessions average 3.62; the full census reads 2.91 | The census (8,880) | §6 |
| The misses are readable, not random | 60 of 83 divergent sessions have a predictable pattern with identifiable frame | Divergent set (83) | §7 |

## 1. Introduction

Every customer-care organization faces the same asymmetry between what it hears and what it measures. The global consumer-goods firm in this study handled 8,880 support conversations across an automated chat assistant and recorded voice calls. Only 537 of them, about six percent, ended with a satisfaction rating the customer chose to give; the other 8,343 left a transcript and no ratings. Satisfaction programs are built on the rated minority, treated as the voice of the customer, even though the customers who supply the ratings select themselves into the measurement (there have been few alternatives). A single overall rating is also a compression: it reduces an interaction involving an agent, an outcome, a product, and some amount of customer effort to one holistic verdict. This paper asks whether a large language model, applied as an annotation instrument, can read the transcript and recover the parts an overall rating compresses, producing a satisfaction score for each component of the experience. It then asks whether, once those annotations are validated against what customers say about themselves, reading them on every contact rather than the self-selected few reveals something the survey cannot.

The annotations were produced by a single large language model (GPT-4.1) applied as a recognition instrument: it read one conversation's text at a time and returned a structured set of satisfaction scores for that conversation alone. For each session the model assigned an overall satisfaction score and four component scores (satisfaction with the agent's handling, with the outcome, and with the product, plus the customer's effort). Throughout the paper we call this instrument the LLM annotation, and the five-axis structure it emits the decomposition. We validated the annotations against two forms of customer self-report: an in-chat overall rating customers gave at the close of a conversation, and a longer post-contact survey battery with item-level ratings. We then analyzed the annotations across all 8,880 conversations, surveyed and unsurveyed alike.

Decomposition, breaking satisfaction into component parts, demands more than showing the model can score different categories of satisfaction. When an instrument emits several related scores at once, the question is not only whether each score is accurate but whether the scores measure separate things or one thing under several names. That is a construct-validity question, and it sets the structure of the paper: a convergent test of whether each axis tracks the self-report it should, a discriminant test of whether the axes are distinguishable from one another, and only then a reading of the annotations on the population the survey never reached. The order matters. Nothing the census read reveals can be trusted until the first two tests say what the numbers mean.

Because the paper moves among several populations, we name them once here and hold the names fixed. The census is all 8,880 sessions, of which 537 are surveyed and 8,343 unsurveyed. The in-chat base is the 432 sessions with a native in-chat rating, and it anchors the convergent tests. The survey battery is the 92 cases with item-level survey ratings, and it anchors the dimension-specific tests.

The validation is strong for most of the decomposition and weak in one place we identify. Read against the customer's own rating on the in-chat base, predicted satisfaction ratings track at correlations near 0.65. The disagreements do not look like general drift; they concentrate in a small, identifiable set of divergent sessions, and setting those aside lifts the overall correlation to 0.811 (severe divergences excluded) or 0.914 (the full divergent tail excluded). We are deliberate about what this does and does not show. Excluding poorly matched sessions after the fact is not itself a validation, because it does not demonstrate that the annotation can flag those sessions on its own. What it establishes is a clear and specific path to a materially higher correlation, one that a revised prompt built to recognize these divergent cases would still have to prove in an experiment we have not yet run.

Yet the axes that validate well are also highly collinear, and combining them does not predict the customer's rating any better than the predicted overall score alone, so the decomposition buys nothing as a predictor of the verdict. Its value emerges only when the annotations are read across the whole census, where satisfaction is markedly lower than the survey reports, a full-census 2.91 against the surveyed 3.62.

These results raise the questions the rest of the paper answers: if the decomposition does not improve prediction, what is it for; what the divergent sessions were actually measuring; how much of the surveyed-to-unsurveyed gap is the unsurveyed majority's genuinely worse experience; and what a satisfaction measure owes to the population it never sampled.

The paper makes three contributions. The first is methodological: it tests whether a five-part decomposition measures five separate things or one thing five times, using convergent and discriminant validation rather than accuracy against a single label. The second is diagnostic: it reads every badly divergent transcript to show what the model was scoring and what the customer was scoring. The third is substantive: it uses those transcripts to estimate survey nonresponse bias, finding that the customers who never answer are meaningfully less satisfied than those who do (2.91 versus 3.62).

## 2. Measurement is established; decomposition is the open question

That a large language model can read a piece of text and assign it a category as reliably as a trained human is rapidly gaining traction as an established method. On standard annotation tasks, zero-shot models have been shown to match or exceed trained crowd annotators by wide margins, with intercoder agreement far above the crowd-worker baseline (Gilardi, Alizadeh, & Kubli, 2023), and the result has held across languages, domains, and label sets. For a satisfaction score recovered from a transcript this is the relevant precedent: the model is performing recognition, identifying satisfaction signals already present in the customer's own words rather than generating an opinion of its own. However, almost all of this evidence concerns a single construct measured against a single gold standard label, where the only question is accuracy, namely whether the model assigned the category a human would. A satisfaction decomposition asks something the existing literature has not yet addressed.

When a model emits not one score but several related ones (overall satisfaction, satisfaction with the agent, with the outcome, with the product), accuracy against a single label cannot tell you whether those scores measure several things or one thing measured several times. A model can achieve high agreement with a human coder while tracking a background concept other than the one the codebook specified (Halterman & Keith, 2024), which makes agreement necessary but not sufficient evidence that a label means what its name claims. The well-established discipline that addresses this is not accuracy but construct validity, and its central tools are convergent and discriminant validation: a measure earns its name by correlating with what it should and correlating less with what it should not, assessed across multiple traits and methods (Campbell & Fiske, 1959). A decomposition is a multitrait claim, and so studying it requires showing not only that each axis tracks its own criterion but that the axes are separable from one another, which is a higher (and more informative) bar than correlation alone.

There is good reason to expect customer experience to have separable components rather than a single underlying tone. Our intuition tells us that components a support organization cares about (whether the agent handled the contact well, whether the issue was resolved, whether the product was sound) can move independently. We would similarly expect the effort a customer expends to solve a problem to complement or enhance these measures. Research shows customer effort predicts future loyalty and behaves as a distinct driver rather than a restatement of how satisfied they were (Dixon, Freeman, & Toman, 2010). A credible decomposition must therefore treat effort as its own axis, on its own polarity, and the sign of its relationship to satisfaction becomes a test in its own right: if predicted effort were merely dissatisfaction with a different label, it would carry no information the satisfaction axes did not already hold.

The deeper reason a decomposition might be worth the trouble lies in what a single overall rating leaves out. A number on a scale is a holistic verdict[1]: a single overall judgment that integrates the whole experience, including the routine satisfaction a customer feels but never thinks to mention. What a customer writes is not that summary. An open-ended response surfaces what was salient[2], the specific and vivid incidents that come to mind, while the rating integrates them together with everything that went unremarked. The two formats elicit psychologically different things, episodic recall on one side and a global, summary evaluation on the other (Schwarz, 1999). The transcript can retain aspect-level information that the customer's own overall rating compresses away.

A second asymmetry worth noting separates the rating from the contact. The customers who return a rating are not a representative sample of the customers who make contact; respondents differ systematically from nonrespondents, and the direction and size of that difference are rarely observable, because the nonrespondents leave no rating to compare against (Groves, 2006). A measurement read from the transcript is subject to neither limit: it preserves the structure the rating compresses, and it exists for every contact rather than only the ones that answered. What such a measurement reveals once it is trusted, and whether it can be trusted, is the subject of this paper.

## 3. Data, annotation, and validation design

A large consumer-care operation hears from its customers constantly and measures them almost never. Across the study window, a multi-category global CPG leader with an expansive health and nutrition portfolio handled 8,880 distinct customer-care sessions across two channels for a single sub-brand over a 30-day period: an automated chat assistant and recorded human-voice calls. Only 537 of those sessions, about six percent, carried a satisfaction rating the customer chose to give; the other 8,343 left a transcript and no score. The rarity of these self-reported scores dramatically increases their value to this study, and demonstrates the need for a predictive scoring solution that can cover a wider swath of customer data. The LLM annotation we validate in this paper reads the transcript, so it produces a satisfaction measurement for every session. This section describes what was measured, how it was produced, and the bases against which we checked it.

*The unit and the corpus.* The unit of analysis is the session: one complete customer-care contact, with incoming turns from the consumer and outgoing turns from a human agent or the automated assistant. The decomposition table contained 8,902 rows covering 8,880 distinct sessions; 22 duplicate rows were resolved by keeping one canonical row per session, leaving 8,880. Two channels are present, an automated chat assistant and recorded voice calls, and they behave differently: voice sessions are longer and carry richer emotional arcs, while chat sessions are shorter and more transactional. Where a session links to a case-management record, structured case metadata (issue reason, severity, risk flags, channel, resolution, handling time) is available; 983 sessions carried such a link.

*Recognition, not generation.* The annotation was executed using the Dimension Labs causal intelligence platform that organizes unstructured customer text into a common schema and applies a large language model to recognize and label information already present in each session. An AI agent is then used to interpret the results. The distinction matters for how the resulting numbers should be read. The model did not author a verdict about the customer; it recognized satisfaction-bearing language in the transcript and mapped it onto a fixed scale; with each session processed independently. The model received one session’s text and returned a structured set of labeled fields for that session alone, with no memory of other sessions. This session-level independence ensures that each label reflects only the language in a single transcript, which allows the scores to be comparable across the entire corpus. The annotations were produced with GPT-4.1 in June 2026.

[1] We use **verdict** to denote the single, holistic overall rating a customer gives: an overall evaluative judgment that integrates the entire experience, including the parts the customer never puts into words, as distinct from the component axes that decompose it.

[2] We use **salience** to denote what a customer emphasizes in their own words: the moments that were most memorable, emotionally intense, unusual, or actionable.

*What the prompt function emits.* A single prompt function decomposes a blended satisfaction signal into eight fields per session. Four are component satisfaction axes on a 1–5 scale: `sat_overall_csat`, the overall predicted satisfaction with the session; `sat_agent_csat`, satisfaction with the agent or assistant’s handling judged independently of the underlying problem; `sat_outcome_csat`, satisfaction with whether the issue was resolved; and `sat_product_csat`, satisfaction with the product or service itself. Two are experience signals: `sat_predicted_effort`, how hard the customer had to work, on a 1–5 scale whose polarity is inverted so that 5 is the most effort and the worst experience; and `sat_recovery_signal`, a five-level categorical trajectory of sentiment across the session. Two are attribution fields: `sat_driver_primary`, the single dominant driver of where satisfaction landed, and `sat_driver_primary_confidence`, the model’s coherence judgment on that attribution. The component axes are scored independently and are allowed to diverge; that divergence, as Section 6 argues, is the analytical payload of the decomposition. In prose we call these the overall, agent, outcome, product, and effort axes; tables retain the field names. Table 1 gives the full attribute set with its source and definition.

**Table 1.** The measured attributes and their sources.

| Attribute | Source | Definition |
|---|---|---|
| `sat_overall_csat` | LLM-annotated | Predicted overall satisfaction with the session (1–5). |
| `sat_agent_csat` | LLM-annotated | Predicted satisfaction with agent or assistant handling, isolated from the product problem (1–5). |
| `sat_outcome_csat` | LLM-annotated | Predicted satisfaction with the resolution, isolated from agent demeanor (1–5). |
| `sat_product_csat` | LLM-annotated | Predicted satisfaction with the product or service itself (1–5). |
| `sat_predicted_effort` | LLM-annotated | Predicted customer effort (1–5; 5 = most effort, the worst experience). |
| `sat_recovery_signal` | LLM-annotated | Satisfaction trajectory across the session (five-level categorical). |
| `sat_driver_primary` | LLM-annotated | Single dominant driver of (dis)satisfaction (seven-level categorical). |
| `CSAT_Score` | Self-reported | In-chat overall satisfaction given at the close of the conversation (1–5). |
| `Overall_Rating` | Self-reported | Survey overall satisfaction (1–5). |
| `Solution_Provided` | Self-reported | Whether the issue was solved (1–5). |
| `Ease_of_Interaction` | Self-reported | Ease-of-interaction (effort) proxy (1–5; 5 = easiest). |
| `Time_Taken_Rating` | Self-reported | Speed of handling (1–5). |
| `Product_Recommendation` | Self-reported | Recommendation intent (1–10; rescaled to 1–5 before comparison). |

**The three bases for comparison.** Validity tests need a ground-truth base and discovery needs a population, and the two are not the same data. We use three, and Table 2 fixes their names and sizes; the paper refers to them by these names throughout. The first is the in-chat base: native overall satisfaction ratings (`CSAT_Score`, 1–5) collected inside the chat channel and linked directly to the session, yielding 432 matched sessions, with usable counts varying by axis because of differing null coverage (from 292 for `sat_product_csat` to 406 for `sat_predicted_effort`). The second is the survey battery: a multi-item post-contact survey at case grain carrying an overall rating (`Overall_Rating`), a resolution item (`Solution_Provided`), an ease item (`Ease_of_Interaction`), a speed item (`Time_Taken_Rating`), and two product proxies (`Product_Recommendation`, 1–10), which links to 105 sessions across 92 cases and supplies the only sub-item ratings, making it the base for dimension-specific convergent tests. The third is the census: all 8,880 sessions, of which 537 carry some validated self-report and 8,343 carry none. The two validation bases do not overlap, and their union is exact: the 432 in-chat sessions plus the 105 battery-linked sessions are the 537 surveyed sessions of the census. The first two bases answer whether the annotation works; the third is where it is used. Two derived cohorts appear later and are also defined in Table 2: the collinearity sample, the 3,310 census sessions where all five axes are non-null, used for the inter-axis correlations in Section 5; and the divergent set, the 83 in-chat sessions where the annotation and the customer disagree by two points or more, read one by one in Section 7. Figure 1 shows how the cohorts nest.

**Table 2.** The analytic cohorts. Every statistic in the paper names the cohort it was computed on.

| Cohort | Size | Definition |
|---|---|---|
| The census | 8,880 | All distinct sessions after deduplication (from 8,902 rows); the population where the validated annotation is used (Section 6). |
| The in-chat base | 432 | Sessions with a native in-chat CSAT_Score; usable n varies by axis from 292 to 406. Anchors the convergent tests (Section 4). |
| The survey battery | 92 cases | Cases with a linked multi-item survey (105 sessions before collapse); usable n varies by pairing from 72 to 87. Anchors the dimension-specific tests (Section 4). |
| The census — surveyed | 537 | Sessions with any validated self-report. |
| The census — unsurveyed | 8,343 | Sessions with no self-report. |
| The case-linked subset | 983 | Sessions with case metadata. |
| The collinearity sample | 3,310 | Census sessions where all five axes are non-null; used for the inter-axis correlations (Section 5). |
| The divergent set | 83 | In-chat sessions where the predicted overall score and the customer differ by two points or more; read in transcript in Section 7. Adding the stitched survey lane brings the combined count to 88. |

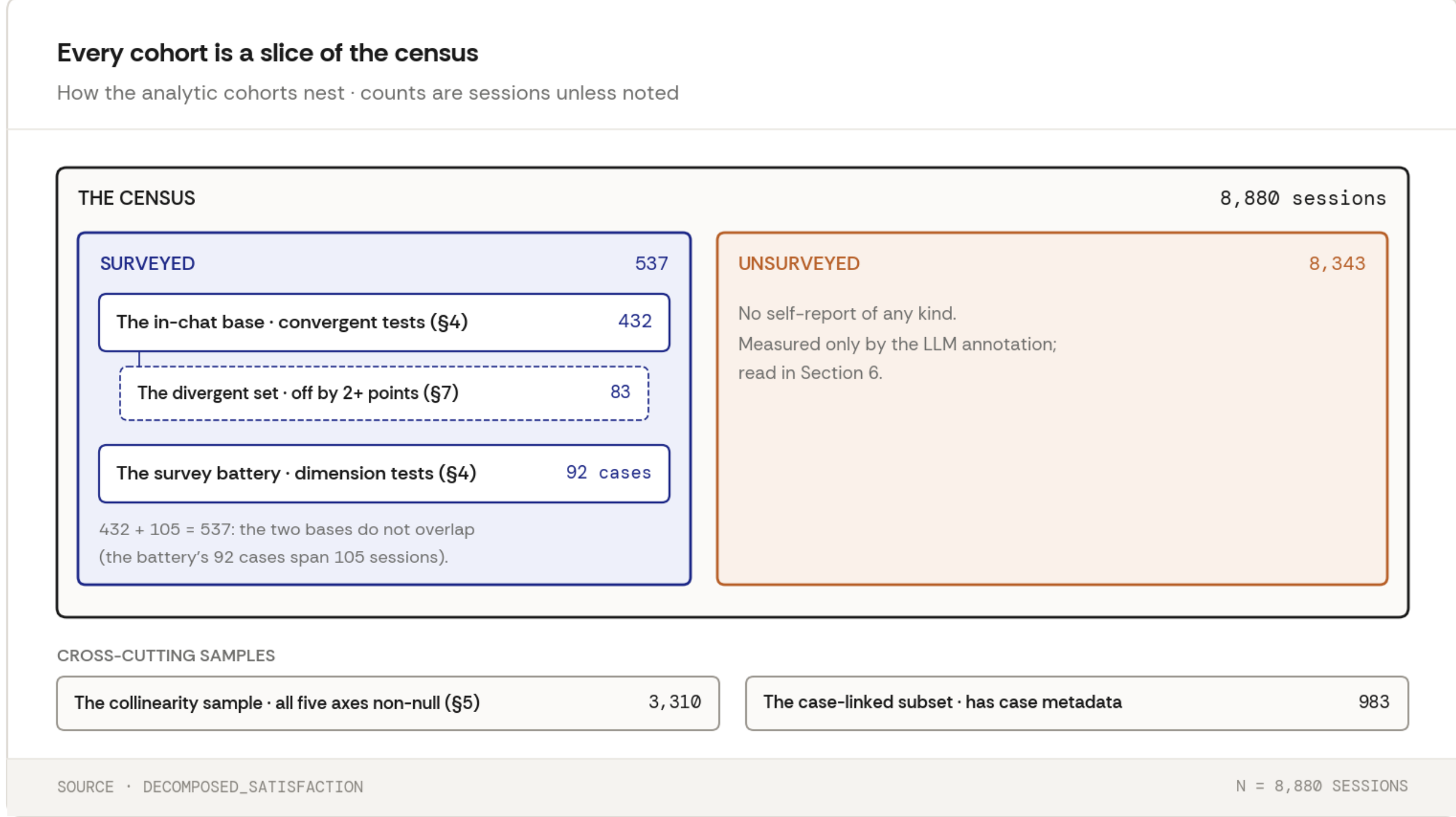

**Figure 1.** How the analytic cohorts nest. The two validation bases on the surveyed side do not overlap (432 + 105 = 537), and the divergent set sits inside the in-chat base. The collinearity sample and the case-linked subset cut across the surveyed and unsurveyed populations.

*Linkage and unit-matching rules, fixed before testing.* The LLM annotation is per session while the survey and case metadata are per case, so every survey-linked analysis collapses a case to one representative session, the latest by session start, before any test. Eleven of the 92 survey cases had more than one linked session; for each of them, this rule kept the latest session and set the others aside. The product recommendation proxy is an eleven-point item and was rescaled to a 1–5 comparable scale before any comparison with a predicted score. Predicted effort was kept in its native polarity, with direction aligned explicitly wherever it entered a joint analysis. Nulls were treated as missing and excluded pairwise, never recoded to a category or a zero, because a null component score means the axis was not evaluable in that session rather than that it scored low.

*How alignment was measured.* Because every satisfaction field is ordinal, we report Spearman rank correlations as the primary alignment statistic, with exact-match and within-±1 agreement rates alongside, since a correlation and an agreement rate can tell different stories about the same pair. Uncertainty is reported as analytic Fisher-z 95% confidence intervals. Where a family of confirmatory tests is reported together, we apply both Bonferroni and Benjamini-Hochberg corrections and state which conclusions survive each. Where two correlations share a variable and the question is whether one is reliably larger than the other, we use tests for dependent correlations rather than eyeballing the gap. One planned check was not completed and bounds what follows: the prompt function was not re-run to estimate run-to-run reliability, so we report no reliability ceiling and make no claim about the stability of an individual label across repeated annotation. We return to this in Section 10. The two sections that follow apply the logic of convergent and discriminant validation (Campbell & Fiske, 1959): Section 4 asks whether each axis tracks the criterion it should, and Section 5 asks whether the axes are separable from one another or are one signal scored several times.

## 4. The decomposed scores recover self-reported satisfaction

A customer types four sentences into a chat window and, later, picks a number from one to five. The model reads the same four sentences, never sees the number, and picks its own. The first question of any predictive satisfaction measure is how close those two numbers are, and across the in-chat base (432 sessions) the answer is: close enough to track the customer, on every axis but one.

*The ranking against the customer's own rating.* Correlated against the customer's self-reported overall satisfaction, four of the five annotated axes land in a tight top tier and the fifth behaves exactly as a distinct effort measure should. On the in-chat base, the overall axis correlates at 0.65 ($n = 385$), the outcome axis at 0.64, and the agent axis at 0.64; the product axis trails at 0.56 on its smaller base; and the effort axis runs at −0.54, negative because higher effort means lower satisfaction (per-axis n from 292 to 406; Table B1 gives each). Every one of these relationships clears both Bonferroni and Benjamini-Hochberg correction. As a robustness check, restricting to the 284 in-chat sessions where every axis is scored on the same customers leaves the ordering identical and slightly stronger (0.67, 0.65, 0.66, 0.56, and −0.56 respectively); we use the full-coverage numbers as the headline figures throughout. A reader who expected the purpose-built overall axis to win cleanly will notice that it does not: agent and outcome are within two-hundredths of it. We take that observation up in Section 5.

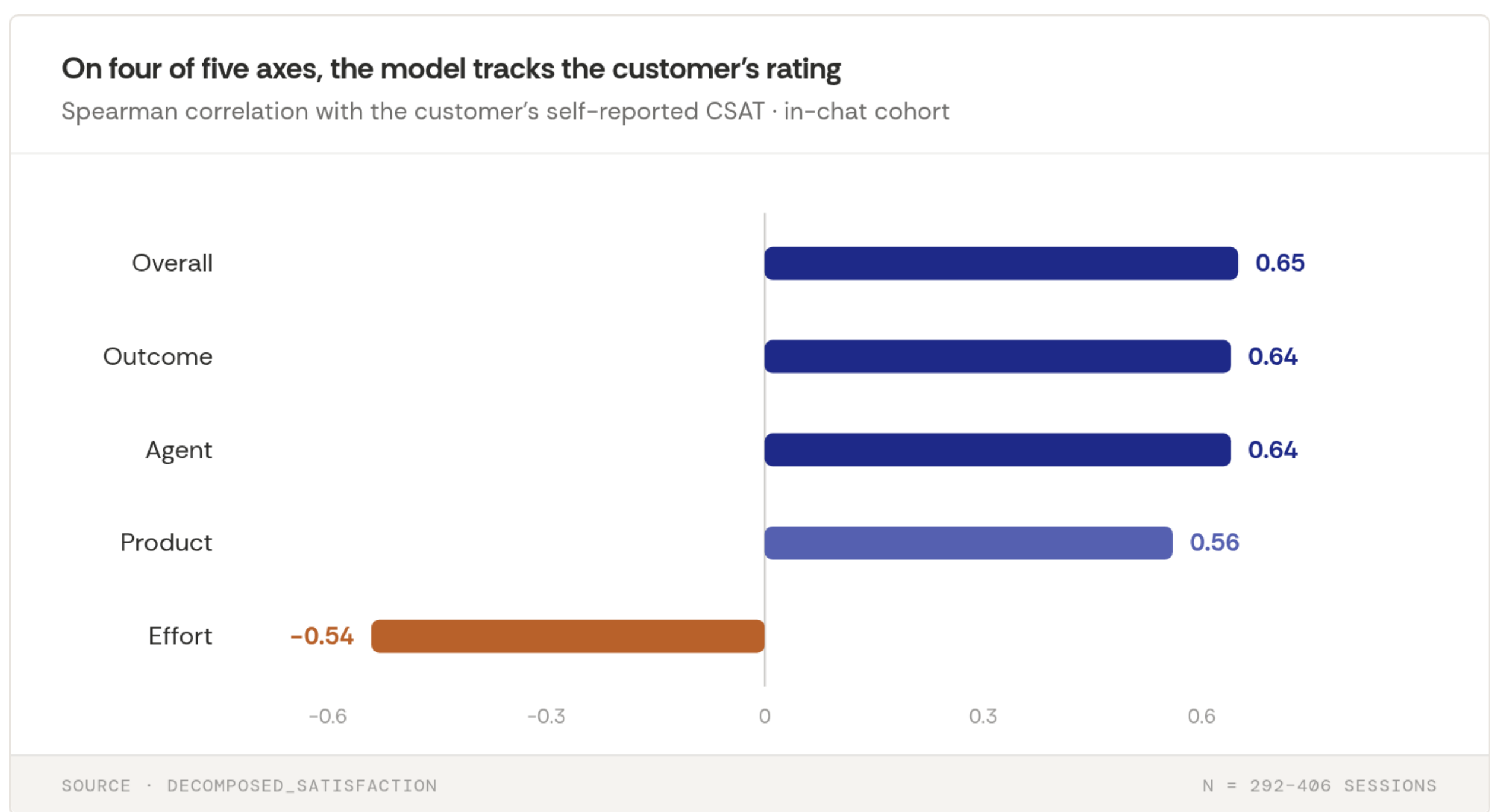

**Figure 2**. Each bar is the Spearman correlation between one predicted axis and the customer's own self-reported satisfaction; longer bars mean tighter tracking. Overall, outcome, and agent cluster near 0.65, product trails at 0.56, and effort is negative (−0.54) because higher effort accompanies lower satisfaction.

**Agreement, not just correlation.** Correlation measures whether the two numbers move together; agreement measures whether they land on the same value. The two come apart. On the survey battery, the overall axis matches the customer's survey overall rating exactly 59% of the time and falls within one point 94% of the time. The outcome axis matches its resolution item exactly 37% of the time but within one point 80% of the time. The exact-match rates look modest precisely because a five-point ordinal scale punishes a one-step miss as harshly as a four-step one; the within-±1 rates are the more honest read of an ordinal instrument, and they are high for the axes that carry the paper's claims. Effort agrees with its ease proxy within one point 80% of the time. Product is the exception on agreement as on everything else, matching its proxy within one point only 38% of the time. High agreement is necessary but not sufficient evidence that an axis measures what its name claims (Halterman & Keith, 2024); a model can land the right number for the wrong reason, which is the question the discriminant tests of Section 5 take up.

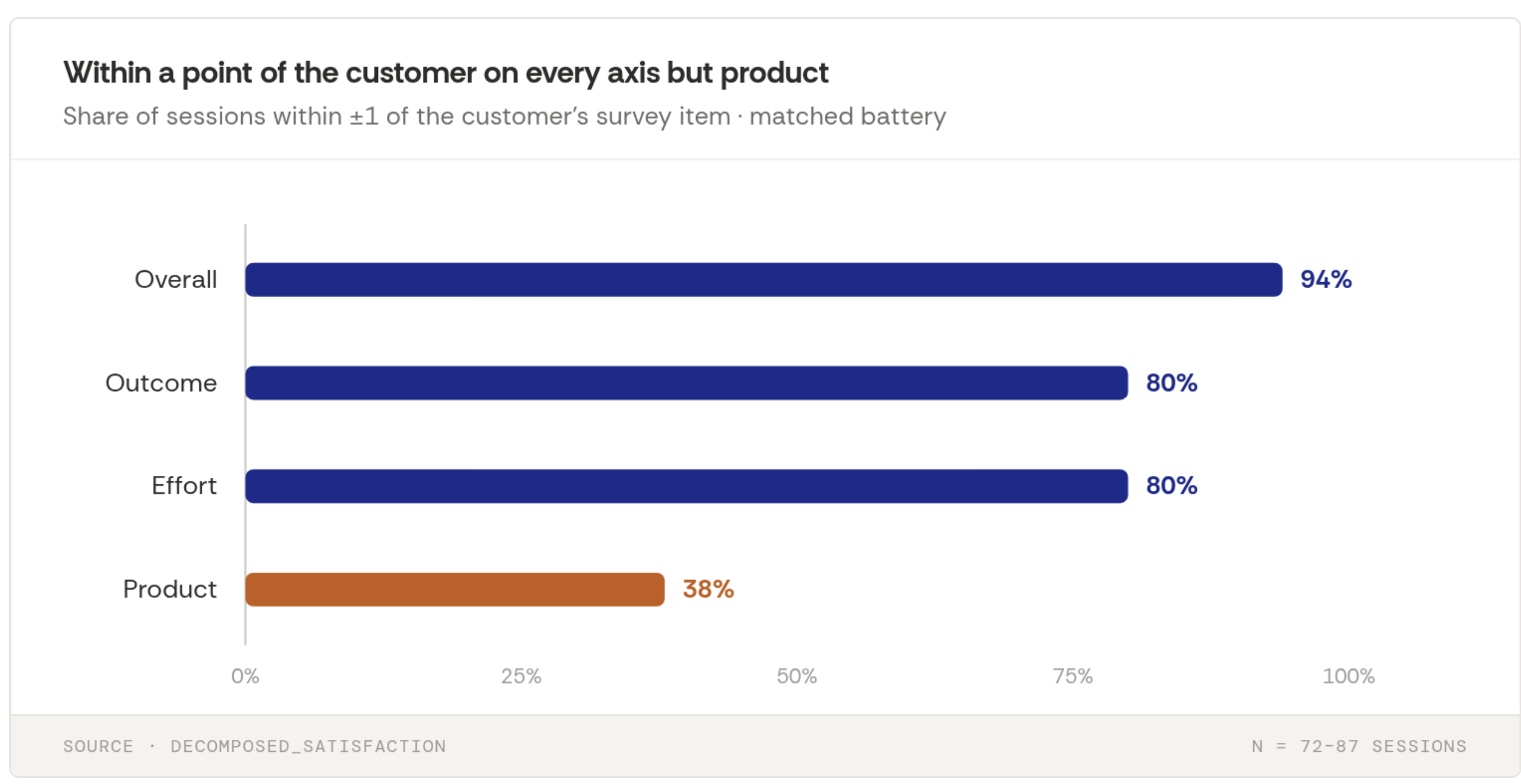

**Figure 3.** Each bar is the share of sessions whose predicted score lands within one point of the customer's matching survey item. Overall, outcome, and effort agree about four times in five; product agrees only 38% of the time.

**Convergent alignment with the matched survey fields.** The in-chat base shows that the axes track the customer's overall verdict; the survey battery (92 cases; per-pairing n from 72 to 87) asks the harder question of whether each axis tracks the specific thing it claims to measure. Here the evidence is positive for three axes and absent for the fourth. The overall axis aligns with the survey overall rating at 0.54 (95% CI [0.37, 0.68]). The outcome axis aligns with the resolution item at 0.45 ([0.26, 0.60]). The effort axis aligns with the ease item at −0.26 ([−0.45, −0.05]) and more strongly with the speed item at −0.40 ([−0.57, −0.21]), the negative sign confirming that the effort axis behaves as a distinct, loyalty-relevant construct rather than a sentiment score wearing a different label (Dixon, Freeman, & Toman, 2010). The battery is small, so these are exploratory and the effort-to-ease relationship survives Benjamini-Hochberg but not the stricter Bonferroni correction. The result that does not appear is the one for product: the product axis correlates with the rescaled recommendation proxy at 0.06 ([−0.17, 0.29]), a relationship indistinguishable from zero, and at −0.06 with the repeat-purchase proxy.

**Table 3.** Each predicted axis against its designated survey field on the matched battery.

| Comparison | n | ρ | 95% CI | Within ±1 | Significance |
|---|---|---|---|---|---|
| `sat_overall_csat` × `Overall_Rating` | 81 | 0.54 | [0.37, 0.68] | 94% | $p < 0.001$ |
| `sat_outcome_csat` × `Solution_Provided` | 86 | 0.45 | [0.26, 0.60] | 80% | $p < 0.001$ |
| `sat_predicted_effort` × `Ease_of_Interaction` | 85 | −0.26 | [−0.45, −0.05] | 80% | $p = 0.02$ |
| `sat_predicted_effort` × `Time_Taken_Rating` | 87 | −0.40 | [−0.57, −0.21] | — | $p < 0.001$ |
| `sat_product_csat` × `Product_Recommendation` | 72 | 0.06 | [−0.17, 0.29] | 38% | n.s. |

*Matched survey battery; n ranges from 72 to 87 per pairing. Spearman ρ with analytic Fisher-z 95% intervals. Overall, outcome, and effort–time survive Bonferroni and Benjamini-Hochberg correction; effort–ease survives Benjamini-Hochberg only; product is non-significant. Product recommendation rescaled from 1–10 to 1–5.*

**The product axis is the soft spot, and it is soft for a reason we can name.** Three axes align with their own survey counterparts; product does not. The most likely explanation is not that the model fails to read product sentiment but that the survey carries no clean product-quality item to check it against. The available proxies measure recommendation intent and repeat-purchase intent, both of which are downstream attitudes shaped by price, loyalty, and habit as much as by the satisfaction expressed in a single support contact. A null against a noisy proxy is weaker evidence of a broken axis than a null against a clean one. We therefore treat product as provisional rather than refuted, and Section 6 shows it behaving sensibly where we can observe it directly, on product-complaint contacts, even though it cannot be confirmed against the survey here.

## 5. One signal or four? Overlap and discriminant validity

A decomposition makes an implicit promise: that four axes carry more than one. The cleanest way to test that promise is also the most demanding. If the component axes genuinely measure separable things, then knowing all four should predict the customer's overall rating better than knowing the overall axis alone. When we ran that test, they did not. Why the extra axes add nothing to the prediction is the question worth answering, and the answer is the real contribution of this paper.

*The composite adds nothing the overall axis did not already carry.* On the 284-session pairwise-complete cohort of the in-chat base, where every axis is scored on the same customers, the predicted overall axis tracks the customer's self-reported satisfaction at 0.67. A simple composite that adds agent, outcome, product, and inverse effort to that overall axis tracks it at 0.65, and a reduced composite dropping the weak product axis reaches 0.66. Adding the components does not improve on the overall score; it very slightly degrades it. The decomposition, judged purely as a predictor of the customer's single rating, is decorative. On this evidence, the component axes do not jointly explain the overall rating beyond what the overall axis explains by itself.

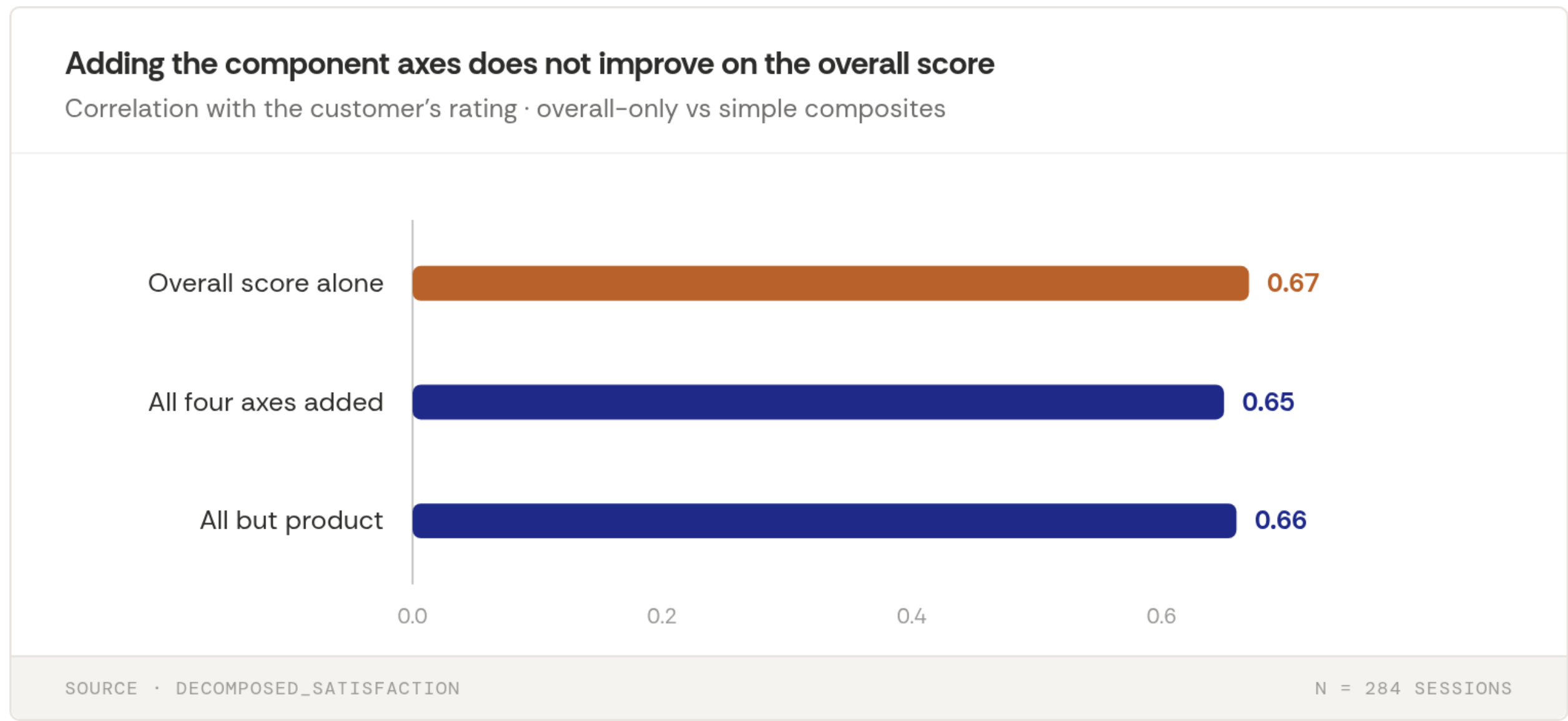

**Figure 4.** Each bar is how well a simple summary predicts the customer's rating: the predicted overall score alone, that score plus all four component axes, and the composite with product removed. The three land within 0.02 of one another, so the components add no predictive power.

**The axes are nearly collinear.** The reason is visible in how the axes relate to one another. On the collinearity sample (the 3,310 census sessions where all five axes are non-null), the overall axis correlates with the outcome axis at 0.94, with the agent axis at 0.85, and with the effort axis at −0.84; agent and outcome correlate at 0.84. The axes do not move independently. A session the model reads as broadly satisfying tends to score high on every axis at once, and a session it reads as a failure tends to score low across the board. Some of this is the world (a contact that goes well tends to go well on every dimension) and some of it is the method (a single model annotating one transcript imposes internal consistency on its own outputs), and the design cannot fully separate the two. Either way, the practical consequence is the same: as predictors of the overall rating, the axes are near-substitutes.

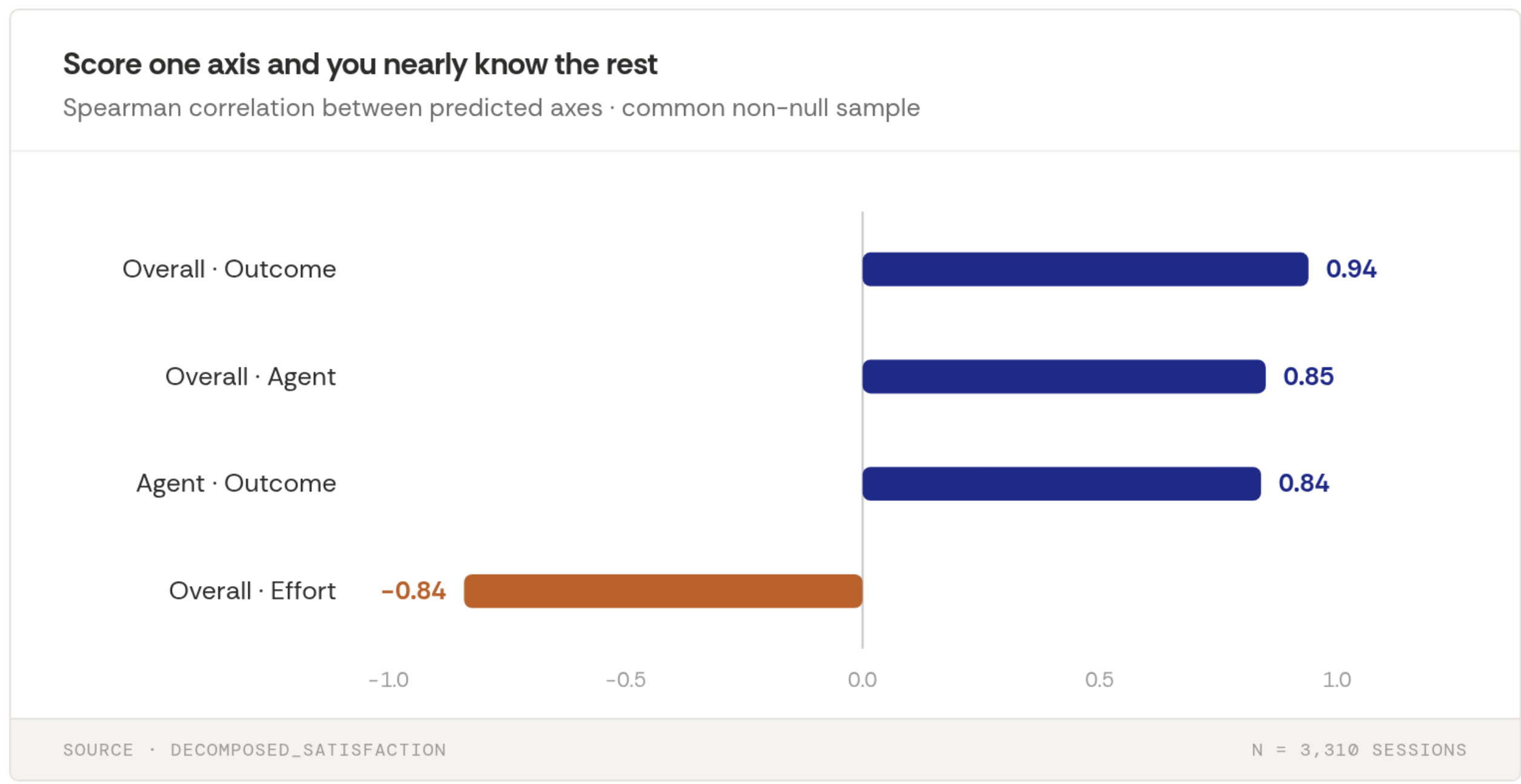

**Figure 5.** Each bar is the correlation between two predicted axes on the sessions where both are scored; the closer to 1, the more the two move together. At 0.84 to 0.94 the axes are near-substitutes, which is why combining them does not improve prediction.

*Redundancy in prediction is not redundancy in meaning.* It would be easy to read this section as bad news, as evidence that the decomposition is one satisfaction signal scored four times. That reading is half right and it misses the point. Two axes can be near-perfect substitutes for predicting a sum and still disagree about its composition on the individual cases where composition is what you need. The discriminant contrasts already show the axes are not collapsing entirely: product and effort hold their distance from the overall verdict even as agent and outcome mirror it. The question the prediction test answers is whether the axes add information about the overall number, and the answer is no. The question Sections 6 and 7 answer is whether the axes add information about why the number is what it is, and the answer is yes. Put plainly, the component axes will not tell you the customer's rating any better than the overall axis already does, but they will tell you what drove it. The value of the decomposition was never going to be incremental prediction of a rating the overall axis already captures. It is **attribution at full coverage**, and that is a different test on a different base.

## 6. What the full census reveals

Everything to this point has been measured on the sessions that carry a rating: the 432-session in-chat base and the 92-case survey battery. The annotation's reason for existing is that it also measures the 8,343 census sessions that carry none, and the moment we look at them the picture of customer satisfaction changes.

*The surveyed minority is the satisfied minority.* The customers who returned a rating are not a smaller copy of the customers who did not. Across the census, the 537 surveyed sessions average 3.62 on predicted overall satisfaction; the 8,343 unsurveyed sessions average 2.91. The gap runs the same direction on every axis we can read: surveyed sessions average 4.00 on agent handling against the unsurveyed 3.42, and 2.43 on effort against the unsurveyed 3.00, meaning the people who answered the survey not only ended happier but worked less hard to get there. This is the nonresponse bias every survey program carries and few can quantify (Groves, 2006), because quantifying it requires a satisfaction measurement for the people who never answered, which is exactly what a transcript-based annotation supplies. The annotation's first contribution to the census is to put a number on what the survey cannot see about itself: read on its own respondents, this operation looks like a 3.6 out of 5; read on its whole population, it looks like a 2.9. The 0.71-point gap between the surveyed and the unsurveyed is the measure of how far a self-selected sample sits from the population it stands in for.

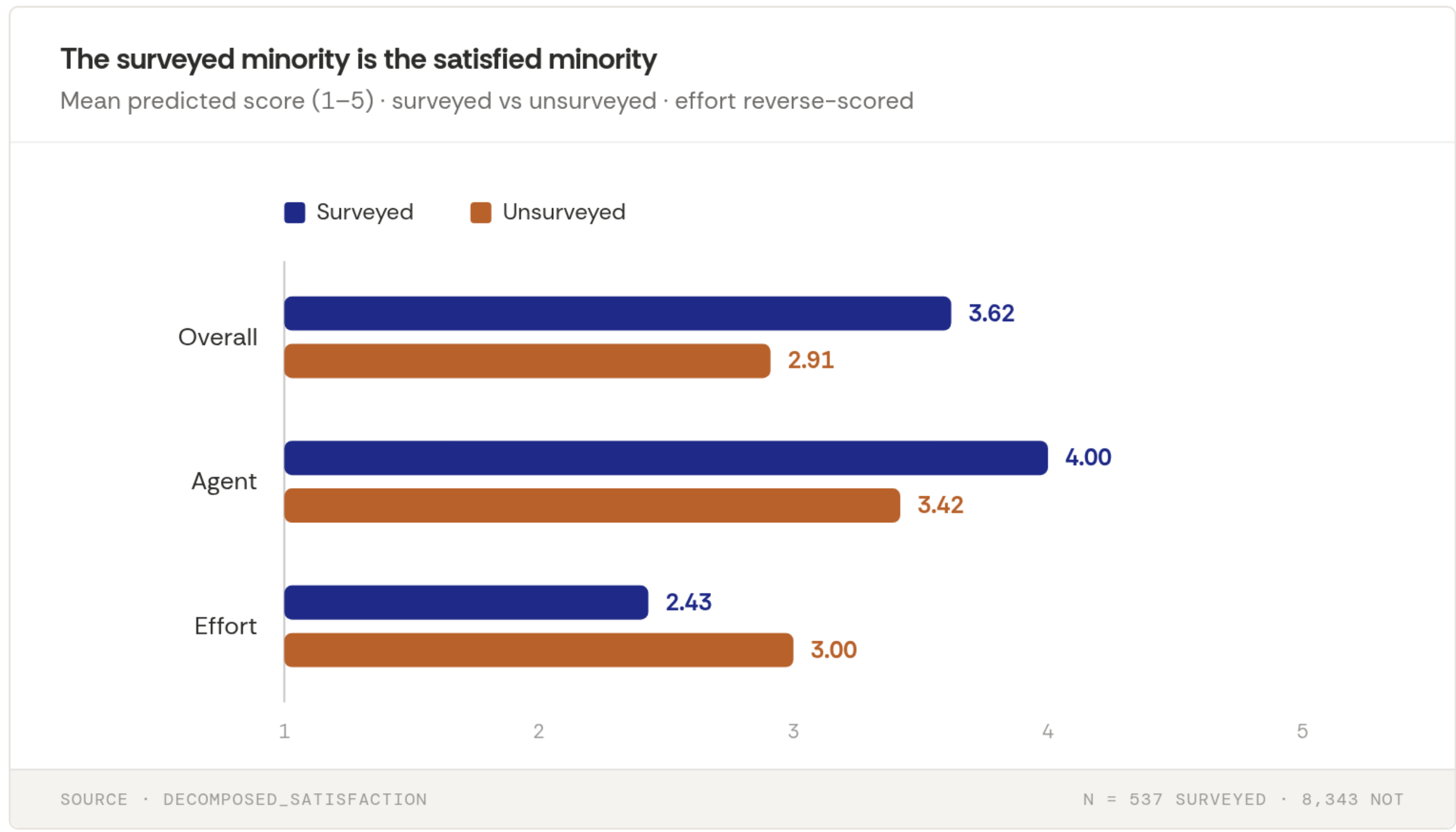

**Figure 6.** For each axis, the two bars compare the mean predicted score among customers who returned a survey (navy) and those who did not (copper). The unsurveyed majority is worse off on every axis; because effort is reverse-scored, its longer copper bar also marks a worse experience.

**One caution travels with this finding.** Validation was only possible where a rating exists, and the customers who rate skew satisfied: the 537 surveyed sessions average 3.62 on predicted overall satisfaction, while the 8,343 unsurveyed sit at 2.91. The annotation is therefore checked at the top of the scale and assumed at the bottom. The direction of the surveyed-to-unsurveyed gap is not in doubt, and the effort and agent axes move with it as they should; but the exact magnitude of the gap inherits whatever error the annotation carries in the low range where it was never directly verified.

*Attribution is what the decomposition buys.* Section 5 showed the axes are near-substitutes for predicting the overall rating, which is why they are read here for a different purpose. The primary attribution is direct: alongside the axes the model records a single dominant driver for each session. Filtered to the high-confidence attributions that cover 8,144 of the 8,880 census sessions (92%), those labels form a map of what drives satisfaction, resolution (2,035 sessions, 25%), the channel or automation experience itself (1,390, 17%), agent handling (914, 11%), product quality (669, 8%), and effort or friction (416, 5%), with policy or compensation rare (93, 1%) and a large neutral remainder (2,627, 32%). This map holds for every contact rather than the few that returned a survey, and it does not rest on the axes being independent; the model names the driver it read whether or not the numeric axes agree.

The axis scores add a second, narrower signal on top of that label. Because the axes usually move together, the sessions where they pull apart are informative in themselves: in 4.8% of sessions the model reads high agent handling alongside low product satisfaction, the signature of a capable agent facing a genuine product failure, and in 3.0% a high overall score sits atop low product satisfaction. These are a minority, which is both their limit and their point, since they are the contacts where a single blended score misleads about the cause. Table 4 shows three sessions where the axes pull apart in exactly this way. What the divergences do not establish is clean separation: part of the axes' agreement is one model keeping its own outputs consistent rather than the world moving in lockstep, so the decomposition earns a practical map of where dissatisfaction concentrates, not a factorization of experience into independent parts.

**Table 4.** Three sessions where the decomposition separates service from product or outcome.

| Pattern | Predicted scores | What the customer wrote |
|---|---|---|
| Capable agent, faulty product | agent 5 · product 1 · overall 4 | *"I've been using [the product] for over 20 years and never had a problem before, but the last purchase made [my pet] very sick … there was a smell that wasn't normal."* |
| Good service, defective packaging | agent 5 · product 2 · overall 4 | *"They don't snap cleanly like they normally do, and so the foil lets gravy come out and it gets all over my hands."* |
| Pleasant contact, unresolved outcome | agent 4 · outcome 2 · effort 4 | *"I ordered [the product] free sample and it's been 7 days, still didn't get my sample or any update … is it just a misleading Ad?"* |

*Predicted axis scores from the decomposition (1–5). Quotes are customer language with brand names and product terms redacted and filler trimmed; wording is otherwise unchanged.*

**The rare patterns are where a blended score misleads.** Two named patterns are worth isolating because they invert what a single score would suggest. **Fragile satisfaction**, a high overall score resting on high customer effort, appears in only 0.4% of sessions, and **genuine service recovery**, a customer who starts dissatisfied and ends satisfied, in only 6.6%. Both are rare, and their rarity is the finding: most sessions do not recover and most high scores are not secretly fragile, so the few that are deserve attention precisely because they are exceptions a blended metric would average away. The same logic applies at the severe end. The nine risk-flagged sessions in the case-linked subset score far below the rest on every axis (overall 2.22 against 3.69, product 1.17 against 2.66, effort 3.89 against 2.63), which is reassuring as a face-validity check and useless as an incidence estimate, because nine is nine. We report it as concordance, not prevalence.

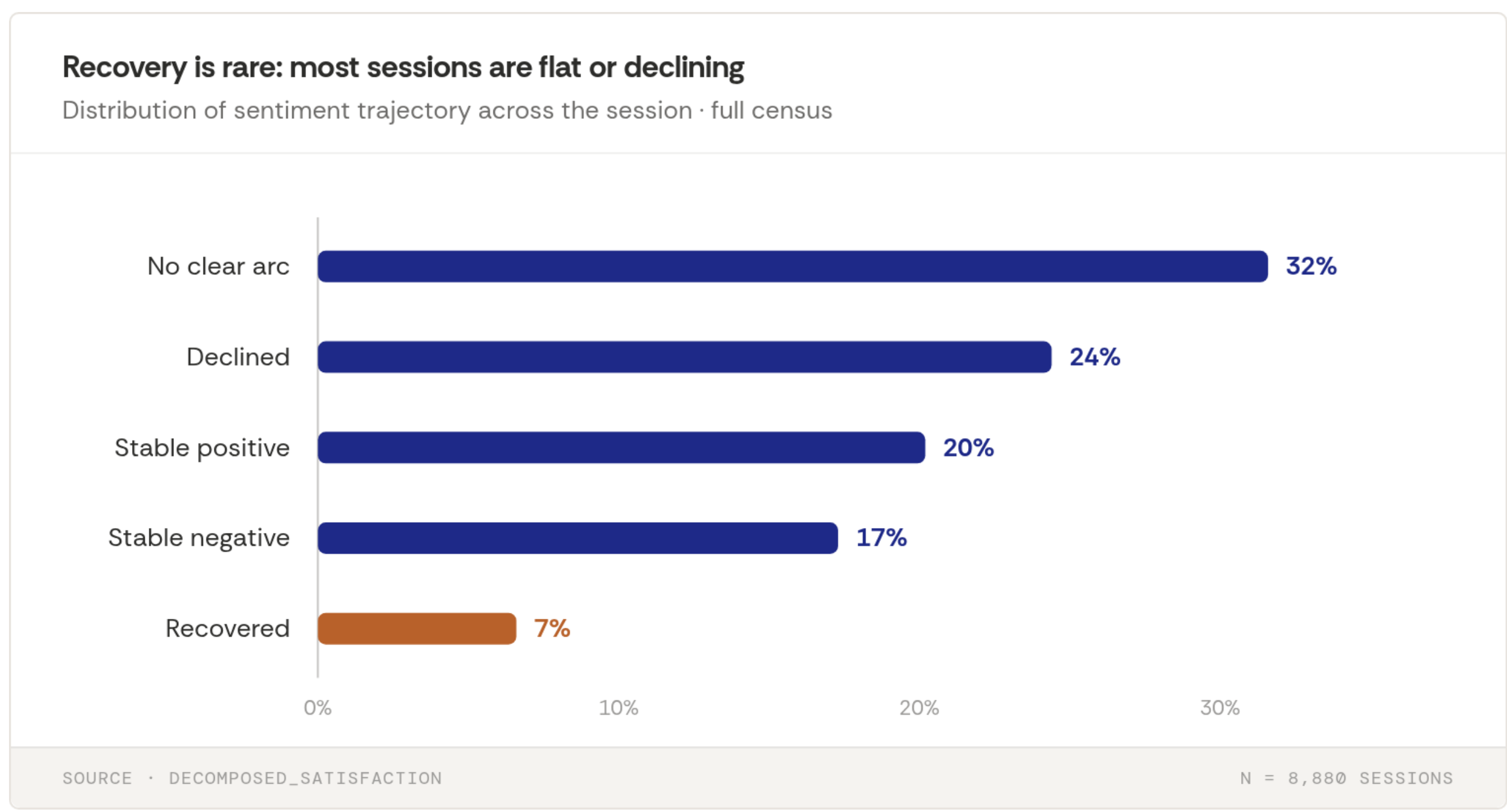

**Figure 7.** Each bar is the share of all census sessions assigned to a sentiment trajectory. Genuine recovery, a session that starts negative and ends positive, is the rarest outcome at 6.6% (charted to whole percents); most sessions show no clear arc or a decline.

**What stays open.** The census read resolves the question it was built for and raises one it cannot close. We can now say, for every contact (rather than the few that answered a survey), where satisfaction landed and what drove it. We cannot yet say how much of the surveyed-to-unsurveyed gap is the unsurveyed majority's genuinely worse experience and how much is the annotation behaving differently outside the range it was validated on, nor can we confirm the product axis against a survey that never cleanly measured product. The decomposition has turned satisfaction from a sample into a census; calibrating that census against the population it now describes is the work this study leaves undone.

## 7. Why the correlation stops at 0.65

A reader arriving at the 0.65 tracking between the predicted overall score and the customer's own rating will ask what the missing distance is made of. It is not general drift. Roughly three-quarters of the in-chat base are exact or near matches, and the shortfall sits in a minority tail: across the combined self-report bases, 88 sessions differ from the customer by two points or more, 35 of them by three or more. Excluding the full divergent tail lifts the coefficient from 0.65 to 0.914; excluding only the 35 severe cases lifts it to 0.811. The limiting factor is a specific, readable subset. So we read it. The in-chat base contributes 83 of the 88 divergent sessions (the divergent set of Table 2; the stitched survey lane adds the other five, all underpredictions), and we read the full transcript of every one. The question is then not how large the misses are but what kind they are.

*They are not, for the most part, a model that cannot read.* Across the in-chat base the annotation carries almost no net bias in level: it sits below the customer a little more often than above (176 sessions to 140, the rest exact), and the average signed error is within a hundredth of zero. What structures the misses is not their direction but their frame. The customer and the model were each asked for one number, and each compressed the same contact along a different axis. Among the 83 sessions of the divergent set the split is 39 overpredictions and 44 underpredictions, and the two have different causes. Overprediction comes from the model scoring the manner of the conversation. Underprediction comes from the model scoring an unresolved outcome the customer did not hold against the contact.

*Overprediction is the more uniform of the two, and the more instructive about what the annotation still lacks.* In 23 of the 39 overpredicting sessions the transcript is fluent and on-brand from open to close. The assistant offers guided options, returns a competently formatted answer, and signs off with a courteous "did I answer your question?" or "what can I help you with next?", while the customer's actual task went only partly done. The model keys on that surface: the satisfaction-bearing language it recognizes is the assistant's own poise, not the customer's resolution. Usually the miss is an intent switch the assistant never makes. A customer inside an informational flow asks "Where can I buy the [product] 1.8 kg.", and the assistant keeps offering guidance instead of a place to buy; predicted overall four, reported one. Another pivots partway in, "teach me how to train my [pet] to sit", and the assistant stays on the prior thread; again four against one. The most extreme cases pair a confident five with a reported one, the model returning a high-confidence positive driver and a stable-positive trajectory over an exchange whose polish it has mistaken for completion. This is consistent enough to name the axis behind it. The decomposition scores whether the issue was resolved, the outcome axis, but never whether the customer's specific job was completed, and a capable-sounding conversation that never finishes that job is exactly where the two part company. The overprediction tail is the fingerprint of a task-completion axis the annotation does not yet carry.

*Underprediction runs the other way, and most of it is recoverable by an axis the annotation already has.* In 29 of the 44 underpredicting sessions the issue is never resolved inside the conversation, and the model, scoring that unresolved outcome, marks the session down while the customer rates it near the top. In the transcripts the customer is plainly scoring the support process. One writes, "My order says it's been delivered but only one item from my order has been delivered, the large bag of [product] is missing"; the assistant cannot fix it in the channel but captures the detail, and the customer rates the contact five against a predicted two. In another, an unmet request, "I applied for a free sample but I haven't received any email regarding that", ends in a clean handoff: "Sent it! Your case number is [number] and you should receive a response within 2 working days." The customer rewards the routing with a five. Escalation requests behave the same way. "Pliz I want to speak to a human" resolves nothing in-chat, yet still earns a top rating for taking the customer somewhere. What makes these more than anecdote is where the other axes land. In these sessions the agent-handling axis sits on or beside the customer's number where the overall axis does not, recovering case by case precisely where the overall axis misses. A further 8 underpredictions turn on product or operational friction the overall score smoothed over: a blocked verification step ("I want free sampal but not verify OTP verification"), or a subscription credit the customer was "unable to" apply. There the product axis, provisional though it is, moves toward the customer.

*The genuine defects are the residual, and they are the minority.* Thirteen of the 83 are real model or assistant failures: an intent misread, a short input the assistant cannot parse, a bad route. Unlike the frame divergences, these are operational bugs rather than measurement artifacts, and the clearest candidates for a fix. The last 10 resist transcript explanation. Six read as survey-side noise, where cooperative or even complimentary language still carries a bottom score: one customer, having called the assistant "very helpful", picks a one and explains only that they "wanted to press something". The other four are too sparse, off-topic, or persona-driven to support satisfaction inference at all. Sixty of the 83, then, are frame divergences with an identifiable cause; only about a quarter are defect or noise. Table 5 summarizes the full taxonomy, and Table 6 shows representative sessions.

**Table 5.** The six failure modes across the divergent set.

| Failure mode | Direction | Sessions | Share |
|---|---|---|---|
| Process success or clean handoff undervalued: the customer rewards being heard, a credible next step, or an escalation the model marks down as unresolved | Under | 29 | 34.9% |
| Helpful tone overvalued: the conversation is polished but the customer's task is not completed | Over | 23 | 27.7% |
| Intent, routing, or comprehension failure by the assistant | 10 over · 3 under | 13 | 15.7% |
| Product or operational friction the overall score smoothed over | Under | 8 | 9.6% |
| Rating-frame mismatch or survey-side noise | 4 over · 2 under | 6 | 7.2% |
| Too sparse or off-topic to support inference | 2 over · 2 under | 4 | 4.8% |
| **Total** | **39 over · 44 under** | **83** | **100%** |

All 83 transcript-reviewed sessions of the divergent set (in-chat base); shares are of the 83. The two frame-divergence modes at the top account for 62.7% of the set.

*Two things follow, and both cut in the decomposition's favor.* The first: the disagreements are evidence for the annotation, not against it. Where the model's single verdict and the customer's single verdict pull apart, the components name what each was tracking, the customer's process satisfaction on the underpredictions and the incomplete task on the overpredictions. That is the attribution a blended score cannot perform, and the reason the correlation ceiling is not the right measure of the annotation's worth. The second is an asymmetry worth stating plainly, because it tells a practitioner where to spend. Underprediction is largely a solved problem hiding inside the existing decomposition: the agent axis already recovers it, and a composite that leans on agent handling in escalation-heavy contacts would close much of the gap. Overprediction is not solved, because the axis that would explain it, whether the customer's task actually completed, is one this decomposition does not measure. Read closely, the misses are less a verdict on the scores than a specification for the next one.

**Table 6.** Where predicted overall and the customer diverge, and what each was scoring.

| Direction | Reported | Predicted overall | Gap | What the customer was scoring | Verbatim |
|---|---|---|---|---|---|
| Over | 1 | 4 | +3 | Task not completed; intent switch ignored | "Where can I buy the [product] 1.8 kg." |
| Over | 1 | 4 | +3 | Task not completed; pivot not followed | "teach me how to train my [pet] to sit" |
| Over | 1 | 5 | +4 | Polish mistaken for completion | guided flow, continuation cues: "Got it! What comes next?" |
| Over | 1 | 3 | +2 | Direct question left unanswered | "Where can I find your recycling information please" |
| Under | 5 | 2 | −3 | Support process; detail captured for follow-up | "…only one item from my order has been delivered, the large bag of [product] is missing" |
| Under | 5 | 2 | −3 | Support process; clean handoff with a case number | "I applied for a free sample but I haven't received any email…" → "Sent it! Your case number is [number]…" |
| Under | 5 | 2 | −3 | Support process; routed to a human | "Pliz I want to speak to a human" |

| Direction | Reported | Predicted overall | Gap | What the customer was scoring | Verbatim |
|---|---|---|---|---|---|
| Under | 5 | 2 | −3 | Operational friction; verification blocked | “I want free sampal but not verify OTP verification” |
| Noise | 1 | 5 | +4 | Transcript contradicts the score | “[they] were very helpful, I just wanted to press something” [trans.] |

*Illustrative sessions from the divergent set (in-chat base); predicted overall and reported CSAT on a 1–5 scale. Customer language with brand and product-specific terms redacted in square brackets and filler trimmed; wording is otherwise unchanged.*

## 8. Discussion and Conclusion

A satisfaction survey measures the customers who choose to answer it. Read from the transcript instead, satisfaction can be scored for every customer who made contact, and in this corpus that population is both larger than the survey's and less satisfied: where the customers who answered average 3.62 on a five-point scale, the unsurveyed majority who did not sits at 2.91. That 0.71-point gap is not noise. It is the nonresponse bias every survey program carries and few can quantify, and quantifying it is the first thing a transcript-based measure offers that a survey cannot recover about itself.

That measure earns the name. Read against the customer's own rating, the predicted overall, agent, and outcome axes track it at Spearman correlations near 0.65 and, on the survey battery, agree with it to within a point 94% of the time; predicted effort runs negative, as a distinct and loyalty-relevant axis should, at −0.54. Only product fails to confirm, and it fails against proxies, recommendation and repeat-purchase intent, too noisy to convict it, so we carry it as provisional rather than refuted. What the annotation recovers is the customer's own verdict, read out of the words the customer wrote rather than manufactured as an opinion of its own.

The 0.65 understates that alignment, and reading the disagreements says why. They do not spread as general drift; they concentrate in a small, legible tail of divergent sessions, and setting that tail aside lifts the correlation to 0.811 and then 0.914. That is headroom, not a validated gain: excluding the hard cases after the fact does not show the annotation could flag them on its own. But the misses are interpretable, and their two directions have different causes. The model overpredicts when it scores the surface of a fluent, courteous exchange whose task was never finished, and underpredicts when it marks down an unresolved outcome the customer forgave because the contact routed them somewhere useful. The first pattern is the one the present axes cannot absorb, and it names the axis this decomposition is missing: whether the customer's specific task actually completed.

Sharper prediction of that rating was never the decomposition's job, and it does not deliver it: the axes are collinear enough, between 0.84 and 0.94, that stacking them onto the overall axis predicts the customer's number no better than the overall axis alone. What they buy instead is attribution. Across the census the model's driver labels map what moves satisfaction, resolution, channel and automation, agent handling, product, and effort, for every contact rather than the few who answered a survey, and on the sessions where the axes split they mark a capable agent apart from a failing product. The value is a practical read of where dissatisfaction concentrates, at full coverage, which is the one thing a single blended score cannot give.

The through-line is a change in what satisfaction measurement can be. A single rating compresses a contact into one number and exists only for the self-selected minority who answer; a validated decomposition breaks that number into its parts and produces it for the whole population. It moves a satisfaction program from a scorecard, a level for the customers who replied, toward a diagnosis, an attributable account of what shaped the experience and for whom, across everyone who made contact.

Two limits set the price of the discovery. The validation lives among the satisfied, because only the satisfied leave ratings to check against, so the direction of the surveyed-to-unsurveyed gap is secure while its exact size in the low range is not. And every figure here is conditional on one model and one prompt, scored once. The Limitations section takes both up in full; neither unsettles the structural finding, that the surveyed are the satisfied and the transcript can measure the rest.

What remains is to raise the agreement between the model's read and the customer's own rating. Closing that distance is achievable: the divergent sessions are not random noise but a few recognizable patterns, and a prompt rebuilt to read them, with a task-completion axis added to the five, would turn the correlation headroom into a validated gain. What this study cannot yet claim is the gain itself, since setting those sessions aside after the fact is not the same as showing a revised prompt can recognize them on its own. Demonstrating that improvement on ratings the prompt has not seen should be the focus of a future study to solidify the reliability of the method.

## 9. Limitations

**One model, one prompt.** Every number here is conditional on a single model and a single prompt design. A different model or a reworded prompt could shift both the levels and the alignment, and nothing in this study bounds that sensitivity.

*No measured reliability ceiling.* We did not re-run the prompt function to estimate run-to-run agreement, so we cannot report how stable an individual label is across repeated annotation. Previous research has found high-stability across outputs of multiple annotation runs for scalar predictions. However, an ideal study would establish a floor on what a perfectly stable version of the same annotation could achieve.

*Validation on a range-restricted cohort.* The self-report bases skew satisfied, which compresses variance and attenuates correlations, and more importantly means the annotation was checked on a happier population than the one the census read describes. The full-census representativeness finding is an out-of-sample extension and is reported as such.

*A small survey battery.* The dimension-specific convergent tests rest on roughly 80 to 90 matched cases, which makes them exploratory and leaves the product-proxy null in particular underpowered to distinguish a weak axis from a noisy criterion.

## Appendix A. Coding schema

The prompt function emits eight fields per session. Component axes are integers 1–5 returned as strings; categorical fields are always populated; satisfaction axes are null when the session offers insufficient signal to score the axis (for example, `sat_agent_csat` is null when no agent or assistant turn is present). A null is not a low score and is excluded pairwise from every analysis.

**Table A1.** The eight decomposition fields, their output types, and scoring rules.

| Field | Output | Definition and scoring |
|---|---|---|
| `sat_overall_csat` | 1–5 or null | Overall predicted satisfaction with the session, weighted toward where the customer lands by the end. |
| `sat_agent_csat` | 1–5 or null | Satisfaction with agent or assistant handling (courtesy, ownership, accuracy), judged from agent turns and isolated from the product problem. |
| `sat_outcome_csat` | 1–5 or null | Satisfaction with the resolution (did the customer get the result they wanted), isolated from agent demeanor. |
| `sat_product_csat` | 1–5 or null | Satisfaction with the product or service itself (quality, value, formulation), isolated from the support interaction. |
| `sat_predicted_effort` | 1–5 or null | Customer effort expended; inverted polarity, 5 = most effort and worst experience. |
| `sat_recovery_signal` | categorical | Sentiment trajectory: improved_recovered, declined_worsened, stable_positive, stable_negative, no_clear_trajectory. |
| `sat_driver_primary` | categorical | Dominant driver: agent_handling, outcome_resolution, product_quality, effort_or_friction, channel_or_automation, policy_or_compensation, none. |
| `sat_driver_primary_confidence` | High/Med/Low | Coherence of the driver attribution; filter to High before reporting a driver distribution. |

## Appendix B. Supplementary tables

**Table B1.** Dimension ranking against in-chat self-reported CSAT (full coverage).

| Axis | n | ρ | 95% CI |
|---|---|---|---|
| sat_overall_csat | 385 | 0.65 | [0.59, 0.71] |
| sat_outcome_csat | 396 | 0.64 | [0.58, 0.69] |
| sat_agent_csat | 401 | 0.64 | [0.58, 0.69] |
| sat_product_csat | 292 | 0.56 | [0.48, 0.64] |
| sat_predicted_effort | 406 | −0.54 | [−0.61, −0.47] |

*All relationships $p < 0.001$; all survive Bonferroni and Benjamini-Hochberg correction.*

**Table B2.** Primary driver distribution. Shares are of the 8,144 census sessions with a high-confidence attribution (92% of the census).

| Primary driver | Sessions | Share |
|---|---|---|
| none (positive, neutral, or transactional) | 2,627 | 32.3% |
| outcome resolution | 2,035 | 25.0% |
| channel or automation | 1,390 | 17.1% |
| agent handling | 914 | 11.2% |
| product quality | 669 | 8.2% |
| effort or friction | 416 | 5.1% |
| policy or compensation | 93 | 1.1% |

**Table B3.** Satisfaction trajectory and divergence-pattern prevalence (full census).

| Pattern | Sessions | Share |
|---|---|---|
| Trajectory: no clear trajectory | 2,802 | 31.6% |
| Trajectory: declined / worsened | 2,167 | 24.4% |
| Trajectory: stable positive | 1,792 | 20.2% |
| Trajectory: stable negative | 1,533 | 17.3% |
| Trajectory: improved / recovered | 586 | 6.6% |
| Divergence: high agent / low product | 427 | 4.8% |
| Divergence: high overall / low product | 267 | 3.0% |
| Divergence: fragile satisfaction (high overall / high effort) | 36 | 0.4% |
| Divergence: high outcome / high effort | 18 | 0.2% |

## Appendix C. Verbatim reference

This appendix collects additional customer verbatims by theme. As in Table 4, brand names are redacted in square brackets and filler is trimmed; wording is otherwise unchanged. Each quote is illustrative: the quantitative claims remain established by the figures and the tables above.

**Table C1.** Additional verbatims by theme.

| Theme | Predicted scores | What the customer wrote |
|---|---|---|
| Product question, answer-quality miss | overall low | *"My question wasn't really answered. I was just given the same information that's on the back of the [product] container. I wanted to know if it was safe to feed [my pet] more than one container per day."* |
| Negative start, later recovered | recovery: improved · product 2 | *"They started throwing up again, and I noticed … a chemical smell from the food."* |
| Agent-led recovery | agent 5 · recovery: improved | *"I appreciate your professionalism. They trained you correctly, so thank you so much, [the agent]."* |
| Surveyed responder, positive | survey response | *"I would just like to say a special thank you to the whole team, especially [the agent], for the wonderful and kind message."* |
| Surveyed responder, unresolved | survey response | *"Service was atrocious and still no resolution."* |
| Low overall experience | overall low | *"I asked for nutritional info and you returned 'how should I store' your food. Totally unhelpful."* |

## Appendix D. Prompt Function

This appendix includes the prompt function used by the model to annotate the decomposed satisfaction scores to individual records.

```
{
  "name": "extract_session_info",
  "description": "Decompose predicted satisfaction for consumer-care sessions.",
  "context": "{company} is a global {cpg category} business; the specific brand referenced in the data is
{brand}. Incoming messages are from customers and outgoing messages are from support agents or chatbots.\n For
all numerical rating/score fields: output a single digit and never a word form, padded zero, or decimal.\n-
When a conditional field should be empty, output null value (not 'none', 'N/A', or blank string).",
  "parameters": {
    "type": "object",
    "properties": {
      "sat_overall_csat": {
        "description": "An estimation of the overall satisfaction of the consumer with this entire interaction
rated on a scale of 1 to 5, where\n 1 = Very Dissatisfied\n 2 = Somewhat Dissatisfied\n 3 = Neither satisfied
nor dissatisfied\n 4 = Somewhat Satisfied\n 5 = Very Satisfied\n This is the overall predicted CSAT for the
session and the composite the component axes decompose. Weigh where the consumer lands by the END of the
session more heavily than where they started – a session that opens angry but closes with genuine thanks rates
4-5; one that closes in continued frustration rates 1-2. Output the rating as a single digit (e.g., 4, not
'four' or '04'). If there is not enough information to confidently assign a rating, output null value.",
        "type": "string"
      },
      "sat_agent_csat": {
        "description": "An estimation of the consumer's satisfaction with the AGENT or BOT handling of this
interaction rated on a scale of 1 to 5, where\n 1 = Very Dissatisfied\n 2 = Somewhat Dissatisfied\n 3 =
Neither satisfied nor dissatisfied\n 4 = Somewhat Satisfied\n 5 = Very Satisfied\n Evaluate the behavior and
competence of the human agent or bot ONLY – courtesy, empathy, responsiveness, accuracy, ownership, follow-
through – independent of the underlying product problem, which may be outside the agent's control. A skilled,
caring agent who could not fix a defective product still rates high here; a rude, dismissive, or unhelpful
agent rates low even when the product was fine. Judge from the agent/bot turns only. Output the rating as a
single digit (e.g., 4, not 'four' or '04'). If there is not enough information to confidently assign a rating,
output null value.",
        "type": "string"
      },
      "sat_outcome_csat": {
        "description": "An estimation of the consumer's satisfaction with the OUTCOME or resolution of this
interaction rated on a scale of 1 to 5, where\n 1 = Very Dissatisfied\n 2 = Somewhat Dissatisfied\n 3 =
Neither satisfied nor dissatisfied\n 4 = Somewhat Satisfied\n 5 = Very Satisfied\n Evaluate only whether the
consumer got the result they wanted – a refund, replacement, voucher, working fix, or satisfying answer –
independent of how pleasant or skilled the agent was. A curt agent who fully resolved the issue rates high
here; a warm agent who left the matter unresolved rates low. Output the rating as a single digit (e.g., 4, not
'four' or '04'). If the session describes no outcome yet (still in progress, abandoned mid-flow, or a
complaint with no resolution signal) or there is not enough information to confidently assign a rating, output
null value.",
        "type": "string"
      },
      "sat_product_csat": {
        "description": "An estimation of the consumer's satisfaction with the {brand} PRODUCT or service
itself rated on a scale of 1 to 5, where\n 1 = Very Dissatisfied\n 2 = Somewhat Dissatisfied\n 3 = Neither
satisfied nor dissatisfied\n 4 = Somewhat Satisfied\n 5 = Very Satisfied\n Evaluate the consumer's sentiment
toward the {products} or other product/service at the center of the contact – quality, taste, value,
formulation, freshness, packaging – independent of the support interaction. Output the rating as a single
digit (e.g., 4, not 'four' or '04'). If the session does not evaluate a {brand}  product (e.g., a pure
billing, website, coupon, or process-only contact) or there is not enough information to confidently assign a
rating, output null value.",
        "type": "string"
      },
      "sat_predicted_effort": {
        "description": "Prediction of score estimating difficulty level the Incoming user is having with
{brand} in resolving their contact on a scale of 1 - 5, with 5 being the most effort. Capture how hard the
consumer had to work to get to a resolution – repeated contacts, having to re-explain their issue, chasing a
promised callback, navigating a bot loop, being transferred. Higher effort means a worse experience. Output
the score as a single digit (e.g., 4, not 'four' or '04'). If there is not enough information to confidently
assign a score, output null value.",
        "type": "string"
      }
    },
    "required": [
      "sat_overall_csat",
      "sat_agent_csat",
      "sat_outcome_csat",
      "sat_product_csat",
      "sat_predicted_effort"
    ]
  }
}
```